%% file: main.tex
\documentclass[10pt,twocolumn,letterpaper]{article}

\usepackage{cvpr}              
\usepackage{multirow}

\input{preamble}
\definecolor{cvprblue}{rgb}{0.21,0.49,0.74}
\usepackage[pagebackref,breaklinks,colorlinks,allcolors=cvprblue]{hyperref}
\newcommand{\fang}[1]{\textcolor[rgb]{0,0,0} {#1}}

\def\paperID{***} 
\def\confName{CVPR}
\def\confYear{2026}

\title{\textit{Too Vivid to Be Real}? Benchmarking and Calibrating Generative Color Fidelity}

\author{Zhengyao Fang\textsuperscript{1}\thanks{Equal contribution.}\ , 
Zexi Jia\textsuperscript{3 *}, 
Yijia Zhong\textsuperscript{4}, 
Pengcheng Luo\textsuperscript{5},  
Jinchao Zhang\textsuperscript{3 \dag }, \\ 
Guangming Lu\textsuperscript{1}, 
Jun Yu\textsuperscript{1}, 
Wenjie Pei\textsuperscript{1,2}\thanks{Co-corresponding authors.}\\
{\textsuperscript{1}Harbin Institute of Technology, Shenzhen,}\\
{\textsuperscript{2}Peng Cheng Laboratory,}{\textsuperscript{3}Independent Researcher,}\\ 
{\textsuperscript{4}College of Computer Science and Aritificial Intelligence, Fudan University,}\\
{\textsuperscript{5}Institute for Artificial Intelligence, Peking University}\\
\vspace{-20pt}
}

\begin{document}
\maketitle
\input{sec/0_abstract}    
\input{sec/1_intro}

\input{sec/2_related}

\input{sec/3_dataset}

\input{sec/4_model}

\input{sec/5_experiment}

\input{sec/6_conclusion}
\input{sec/ack}
{
    \clearpage
    \small
    \bibliographystyle{ieeenat_fullname}
    \bibliography{main}
}

\input{sec/X_suppl}

\end{document}

%% file: sec/0_abstract.tex
\begin{abstract}
Recent advances in text-to-image (T2I) generation have greatly improved visual quality, 
yet producing images that appear visually authentic to real-world photography remains challenging. 
This is partly due to biases in existing evaluation paradigms: 
human ratings and preference-trained metrics often favor visually vivid images with exaggerated saturation and contrast, which make generations often too vivid to be real even when prompted for realistic-style images.
To address this issue, we present \textbf{Color Fidelity Dataset (CFD)} and \textbf{Color Fidelity Metric (CFM)} for objective evaluation of color fidelity in realistic-style generations. 
CFD contains over 1.3M real and synthetic images with ordered levels of color realism, 
while CFM employs a multimodal encoder to learn perceptual color fidelity. 
In addition, we propose a training-free \textbf{Color Fidelity Refinement (CFR)} that adaptively modulates spatial–temporal guidance scale in generation, thereby enhancing color authenticity.
Together, CFD supports CFM for assessment, whose learned attention further guides CFR to refine T2I fidelity, forming a progressive framework for assessing and improving color fidelity in realistic-style T2I generation. The dataset and code are available at \href{https://github.com/ZhengyaoFang/CFM}{https://github.com/ZhengyaoFang/CFM}.\end{abstract}

%% file: sec/1_intro.tex
\section{Introduction}
\label{sec:intro}
\begin{figure}
    \centering
    \includegraphics[width=1\linewidth]{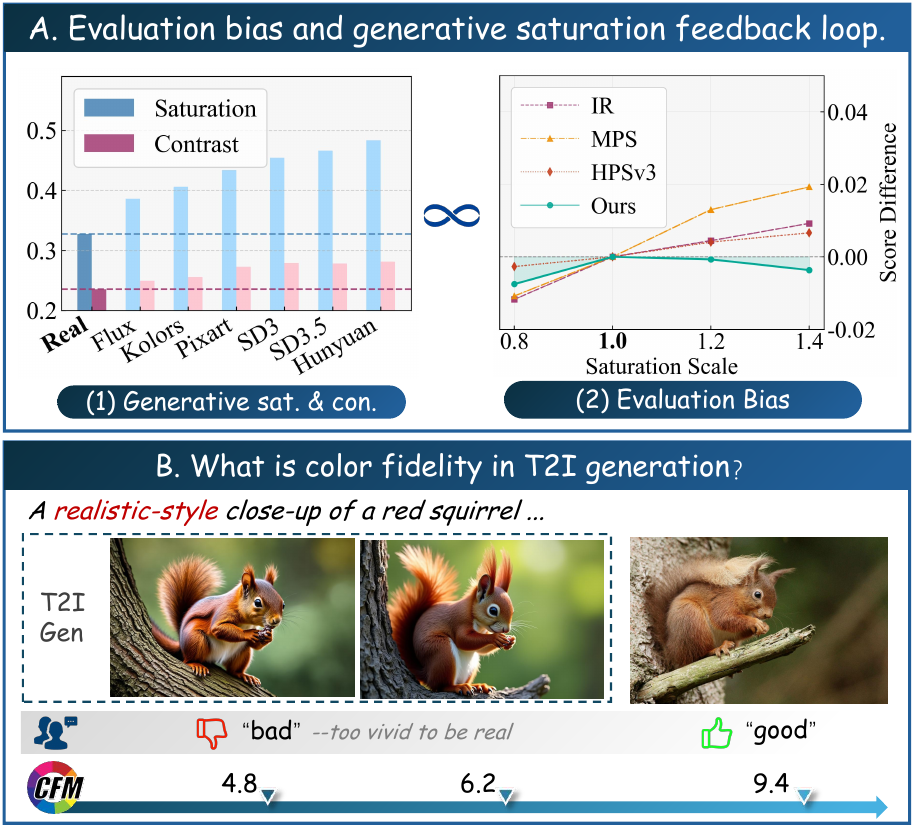}
    \caption{
     \textbf{A. Challenges in existing T2I generation and evaluation.}
     (1) Statistical analysis shows that when prompted to produce \textit{realistic-style} outputs, 
    most T2I models generate images with higher saturation and contrast than real-world photographs.
    (2) A controlled saturation-scaling experiment on real photographs reveals that existing evaluation models exhibit a strong bias toward highly saturated images. The vertical axis shows the normalized score difference with respect to the reference scale (=1.0). 
    \textbf{B. Definition of color fidelity and the target of our CFM.} 
    Color fidelity measures how closely generated images preserve the natural color distribution of real-world photography, 
    which serves as the learning objective of our proposed Color Fidelity Metric (CFM).
    }
    \label{fig:teaser}
    \vspace{-15pt}
\end{figure}

Text-to-image (T2I) generation has made significant progress in recent years, with models such as 
\cite{podell2023sdxl,kolors,chen2024pixart,ding2021cogview,wu2025qwenimagetechnicalreport,li2024playground,esser2024scaling} achieving impressive results in generating 
high-quality images from textual descriptions. However, generating images that appear truly \textit{realistic} remains a challenge. A common issue is color distortion, including over- and under-saturation as well as other photometric artifacts, which compromises the authenticity of the generated images. As illustrated in Fig.~\ref{fig:teaser} A.(1), we perform quantitative analyses on 
\textit{realistic-style} images generated by multiple models using the same set of prompts. 
Compared with \textit{real-world} images, these generated samples exhibit noticeable deviations in 
color fidelity. In particular, most models tend to produce images with higher saturation and contrast, 
a common tendency observed across most models and consistent with the subjective impression of ‘too vivid’ synthetic visuals.

This phenomenon can be partly attributed to biases in existing evaluation paradigms.
Human ratings and preference-trained metrics often favor visually vivid images with exaggerated saturation and contrast,
prompting models to over-optimize for striking colors even when the prompts explicitly request \textit{realistic-style} outputs.
Such biases reveal a fundamental limitation in current evaluation systems: while they excel at assessing semantic relevance or aesthetic appeal,
they \textbf{fail to capture distortions in color realism}.
Metrics such as PickScore~\cite{kirstain2023pick}, ImageReward~\cite{xu2023imagereward}, HPSv3~\cite{ma2025hpsv3}, and
MPS~\cite{zhang2024learning} emphasize semantic alignment or human preference but largely overlook photometric fidelity.
As demonstrated by a controlled saturation-scaling experiment on real photographs (Fig.~\ref{fig:teaser} A.(2)),
these metrics consistently assign higher scores to unnaturally vivid images,
resulting in unreliable evaluations of \textit{realistic-style} generations. As a result, T2I models are implicitly encouraged to generate increasingly saturated images in order to achieve higher scores.

\fang{
This issue stems largely from the absence of a dedicated benchmark and objective metric that explicitly address the dimension of \textit{color fidelity}, as illustrated in Fig.~\ref{fig:teaser}~B.
Here, color fidelity refers to the degree to which a generated image preserves the natural color distribution characteristic of real-world photography, particularly when the prompt explicitly requests a \textit{realistic-style} generation. Measuring such perceptual subtleties requires well-constructed supervision and a modeling framework capable of 
capturing the intricate relationship between color distribution and semantic content. However, 
existing datasets and evaluation protocols offer no systematic way to quantify or learn this 
relationship, making it difficult to evaluate or improve color realism in T2I models.
}

\fang{
To bridge this gap, we introduce the \textit{Color Fidelity Dataset (CFD)} and the corresponding 
\textit{Color Fidelity Metric (CFM)} benchmark. The CFD is a large-scale dataset specifically designed 
to model and quantify color authenticity in realistic-style image generation. It consists of 
high-quality real photographs and corresponding synthetic variants exhibiting progressively distorted 
color fidelity, generated through controlled variation of classifier-free guidance (CFG) 
scales~\cite{ho2022classifier}. This dataset provides explicit supervision of perceptual color realism 
across diverse scenes and categories, enabling systematic training and evaluation of 
fidelity-aware models. Building upon CFD, the CFM employs a vision-language architecture based on 
Qwen2-VL~\cite{wang2024qwen2} to jointly encode textual and visual representations, thereby learning 
a fine-grained perception of color realism beyond global semantics. Trained with a differentiable 
\textit{softrank loss}, the CFM produces consistent color fidelity predictions that correlate strongly with human judgments.
}

\fang{
Furthermore, we extend the benchmark into a practical enhancement framework by introducing a training-free
\textit{Color Fidelity Refinement (CFR)} pipeline. CFR leverages cross-modal 
attention maps from CFM to identify regions exhibiting color--semantic discrepancies and adaptively modulates 
the denoising guidance scale both spatially and temporally. This dynamic adjustment effectively 
suppresses over-saturation and contrast imbalance, yielding more natural and perceptually harmonious 
images without modifying any model parameters or requiring retraining.
}

\noindent
In summary, our contributions are threefold:
\begin{enumerate}[(1)]
    \item we present the \textbf{Color Fidelity Dataset (CFD)}, a large-scale benchmark containing over 1.3M images with explicit supervision of perceptual color authenticity; 
    
    \item we propose the \textbf{Color Fidelity Metric (CFM)}, a multimodal evaluation model trained with ordinal supervision to objectively measure color realism; and
    
    \item we introduce the \textbf{Color Fidelity Refinement (CFR)} module, a training-free, plug-and-play mechanism that improves color realism during generation. 
\end{enumerate}

%% file: sec/2_related.tex
\section{Related Work}
\label{sec:related_work}
\subsection{Text-to-Image Generation}
\fang{
Text-to-image (T2I) generation aims to synthesize visually realistic and semantically aligned images from textual descriptions. With the rapid progress of diffusion-based generative models~\cite{rombach2022high,saharia2022photorealistic,podell2023sdxl,esser2024sd3}, visual autoregressive models~\cite{fan2024fluid,sun2024autoregressive,chang2022maskgit,esser2023structure}, and DiT-based models~\cite{flux2024,li2024hunyuan,peebles2023dit}, T2I generation has achieved remarkable advances in visual fidelity, diversity, and controllability. A key mechanism that drives text–image correspondence in T2I generation is classifier-free guidance (CFG)~\cite{ho2022classifier}, which balances text alignment and visual realism by interpolating between conditional and unconditional predictions during denoising. Subsequent methods have extended this idea through adaptive or learned guidance strategies~\cite{sadat2024eliminating,song2025rethinking}, mitigating the noticeable color distortions that can arise under high guidance scales.}

\fang{Beyond semantic accuracy, generating \textit{realistic-style} images has long been an important goal in T2I research. Recent study~\cite{shen2025directly} has explored enhancing the realism of generated images via direct alignment of diffusion models with human preferences, combining optimized denoising and online text-conditioned reward adjustment. However, despite the growing emphasis on realism, there remains a lack of reliable and quantitative assessment methods to evaluate the authenticity of generated images. Existing evaluation approaches primarily rely on subjective human ratings, which are costly, inconsistent, and difficult to scale, underscoring the need for an objective metric capable of capturing subtle perceptual cues that define visual realism.
}

\begin{figure*}
    \centering
    \includegraphics[width=1\linewidth]{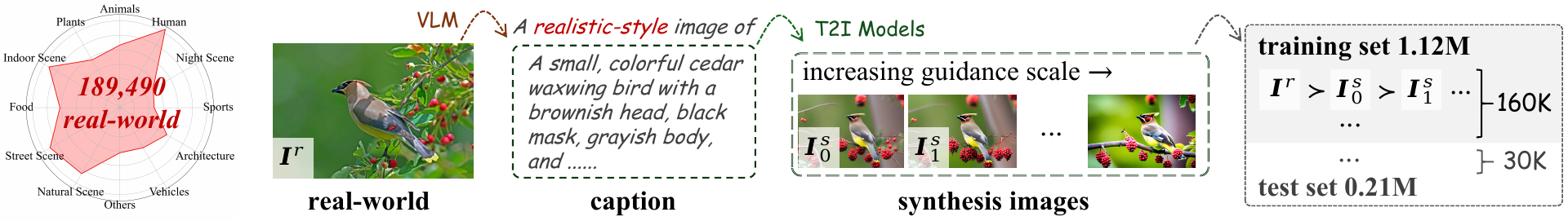}
    \caption{\textbf{Overview of the Color Fidelity Dataset.} 
    We first apply IQA filtering process to obtain about 190K high-quality real-world images across 12 categories. Through \textit{automatic caption generation} and \textit{guidance-controlled image synthesis}, we further construct 1.33M images exhibiting ordered levels of color fidelity, and divide them into training and testing splits.}
    \label{fig:dataset}
    \vspace{-5pt}
\end{figure*}
\subsection{Text-to-Image Assessment}
As discussed in~\cite{hartwig2025survey}, assessing the quality of T2I generation involves multiple dimensions, including text–image alignment, compositionality, realism, aesthetic quality, and creativity. 
Traditional metrics such as FID~\cite{heusel2017gans} and IS~\cite{salimans2016improved} provide useful statistical indicators of overall generative quality, 
yet they operate at the distributional level and thus require a large number of generated samples. 
Consequently, they are unsuitable for evaluating the perceptual quality of individual images or for fine-grained comparisons across prompts and scenes. 
Beyond these limitations, recent research has shifted toward learning-based perceptual assessment methods that better capture semantic and aesthetic aspects. Early metrics such as CLIPScore~\cite{hessel2021clipscore} leverage CLIP similarity to evaluate semantic consistency between text and image. Later works like ImageReward~\cite{xu2023imagereward} and PickScore~\cite{kirstain2023pick} incorporate human preference data to align model evaluation with subjective judgments. More recent efforts, including the HPS series~\cite{wu2023human,wu2023human2,ma2025hpsv3} and MPS~\cite{zhang2024learning}, further refine this approach through large-scale preference datasets and improved multimodal encoders, achieving stronger correlation with human evaluation results. 

However, these approaches predominantly aim to map the overall perceptual quality of a generated image into a single scalar score, often emphasizing global semantics or aesthetic appeal while neglecting specific perceptual factors such as color authenticity and photometric consistency. As a result, they may favor visually striking but perceptually unrealistic images, leading to biased or misleading evaluations, particularly for realistic-style image generation. This limitation highlights the need for more fine-grained and perceptually grounded evaluation paradigms that can accurately capture subtle visual distortions beyond semantic alignment and global aesthetics.

%% file: sec/3_dataset.tex
\section{Color Fidelity Dataset}
\label{sec:dataset}
\fang{
Accurately evaluating the color fidelity of generated images requires a dataset that contains explicit annotations of perceived color fidelity. To this end, we construct a large-scale dataset specifically designed to benchmark \textit{color fidelity} in realistic-style text-to-image generation. Our dataset includes high-quality real-world photographs serving as the upper bound of color fidelity, and corresponding synthetic images exhibiting progressive distortions in color fidelity. This design allows for fine-grained supervision and quantitative assessment of color fidelity across diverse scenes and visual domains. Further implementation details and dataset statistics are provided in the supplementary material.}

\subsection{Preliminaries: Guidance Scale}
\label{sec:cfg}
\fang{A core mechanism of T2I models is the classifier-free guidance (CFG)~\cite{ho2022classifier}. During the denoising process, the model predicts conditional and unconditional noise terms, which are combined as:
\begin{equation}
    \hat{\epsilon}_\theta = \epsilon_\theta(\mathbf{z}_t) + s \cdot (\epsilon_\theta(\mathbf{z}_t, \mathbf{c}) - \epsilon_\theta(\mathbf{z}_t)),
    \label{eqa:cfg}
\end{equation}
where $\mathbf{z}_t$ is the latent variable at timestep $t$, $\mathbf{c}$ denotes the text condition, and $s$ is the \textit{guidance scale}. Increasing $s$ enforces stronger adherence to textual semantics but often introduces over-saturated colors and enhanced contrast, while smaller $s$ values yield faded or washed-out tones. This controllable relationship between $s$ and color statistics provides a principled way to induce and calibrate color distortions. By manipulating the guidance scale, we can generate image sets that exhibit progressively varying degrees of color authenticity while preserving identical semantic content. This makes it ideal for supervised learning of color fidelity.}

\subsection{Dataset Construction}
\noindent\textbf{Real-world image collection.} 
\fang{As shown in Fig.~\ref{fig:dataset}, we begin by collecting a large corpus of high-quality real-world images spanning 12 categories, including humans, natural scenes, and urban environments, which together comprehensively capture the diversity of real-world color distributions. These images are further filtered using an image quality assessment method~\cite{wang2023exploring} to ensure natural color rendering and visual clarity, thereby serving as perceptual upper bounds of color authenticity. In total, 189{,}490 high-quality real-world images are collected for dataset construction.
}

\noindent\textbf{Automatic caption generation.}
\fang{Each image is captioned using the vision-language model~\cite{qwen2.5}, which produces concise and semantically consistent textual prompts for subsequent text-to-image synthesis. These automatically generated captions are designed to preserve both global semantics and fine-grained details, providing reliable prompts for subsequent text-to-image synthesis. 
}

\noindent\textbf{Guidance-controlled image synthesis.}
For each prompt, synthetic images are generated using multiple T2I models with progressively increased guidance scales. Increasing $s$ beyond its default value introduces color distortions that range from subtle enhancement to strong over-saturation while preserving semantics. To ensure comprehensive coverage of color distortion behaviors, we employ 11 T2I models across 12 semantic categories covering both object- and scene-centric content, as models and categories differ in their sensitivity to guidance scales. Each real image is paired with six synthetic variants of increasing distortion, forming a color fidelity sequence of seven images. In total, about 190K groups (1.33M images) are constructed, representing a wide spectrum of realistic-style generations.

\noindent\textbf{Dataset split.}
\fang{The dataset is divided into 160k training and 30k testing groups to support model development and standardized benchmarking. The training split (CFD-Training dataset) is used to learn color fidelity predictors and calibration models, while the test split (CFD-Test dataset) serves as an independent benchmark for evaluating color authenticity in a fair and reproducible manner.
}

\subsection{Human Annotation}
We conduct a comprehensive user study to obtain perceptual color fidelity annotations, where participants evaluate a subset of realistic-style generated images based on visual attributes such as sharpness, illumination, saturation, and overall color realism.
Each image is independently assessed by three trained annotators to ensure inter-rater reliability, resulting in over 20{,}000 human ratings across 6{,}690 images.
These annotations constitute the CFD-Human test set, which provides ground-truth perceptual labels for validating the proposed evaluation framework, and exhibit high inter-rater consistency (average Spearman correlation above 0.85), demonstrating the overall reliability and quality of the collected annotations.

%% file: sec/4_model.tex
\section{Color Fidelity Metric}
\subsection{Architecture}
\label{sec:framework}
\fang{Color fidelity evaluation is closely related to image semantics, as the perception of “realistic color” depends on the contextual and compositional content within the scene. Thus following \cite{ma2025hpsv3}, we employ the Qwen2-VL~\cite{wang2024qwen2} vision-language model as the backbone to jointly encode both visual and textual representations for fidelity-aware feature extraction. Through the Qwen2-VL encoder, the image and text are tokenized and projected into a unified multimodal embedding space. The final token sequence can be expressed as:
\begin{equation}
    \mathbf{F} = [\mathbf{f}_1^{v}, \ldots, \mathbf{f}_M^{v}, \mathbf{f}_1^{t}, \ldots, \mathbf{f}_N^{t}] \in \mathbb{R}^{(M+N) \times d},
\end{equation}
where $\mathbf{f}_i^{v}$ and $\mathbf{f}_j^{t}$ represent the $i$-th visual and $j$-th textual embeddings, respectively, and $d$ denotes the feature dimension. 
This joint representation serves as the foundation for subsequent color fidelity evaluation and refinement.
} 

\begin{figure}
    \centering
    \includegraphics[width=1\linewidth]{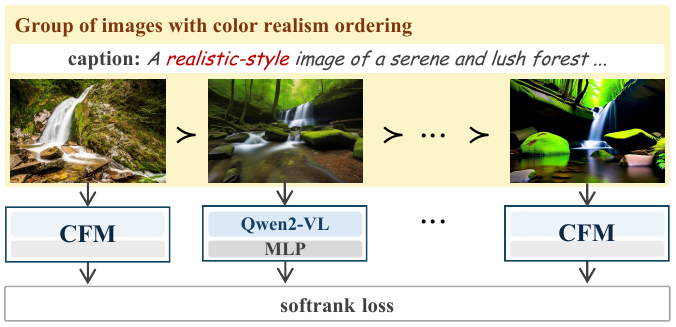}
    \caption{\textbf{Framework of CFM and training pipeline.} CFM employs Qwen2-VL as a multimodal feature backbone and an MLP projection head to map features into scalar fidelity scores. It is trained on the CFD-Training with group-wise, order-preserving samples, optimized by the softrank loss within each group.}
    \label{fig:framework}
    \vspace{-7pt}
\end{figure}
\fang{
To predict the final color fidelity score, the fused feature sequence $\mathbf{F}$ is processed by a multi-layer perceptron (MLP) head, producing token-level logits $\mathbf{z} \in \mathbb{R}^{(M+N)}$. 
Among these, the output corresponding to the special textual token \texttt{<|Reward|>} is selected as the color fidelity score:
\begin{equation}
    S_{\text{CFM}} = z_{\texttt{<|Reward|>}},
\end{equation}
where $S_{\text{CFM}}$ serves as a scalar measure of the perceived color authenticity of the generated image. 
A higher $S_{\text{CFM}}$ indicates stronger alignment between the image’s color distribution and real-world color statistics.
}

\subsection{Training Objective}
\fang{
To train the CFM model, we exploit the inherent ordinal structure of CFD-Training dataset, where each group of $K$ images consists of one real reference and $K-1$ synthetic variants that exhibit progressively decreasing color authenticity as the guidance scale increases. We adopt a differentiable \textit{softrank loss} that enforces consistency between the predicted and ground-truth color realism orderings.
}

\fang{
Specifically, for each group, the model predicts a reward score $r_i$ for each image $\mathbf{I}_i$ ($i=0,\ldots,K-1$). 
All predicted scores are first stacked as:
\begin{equation}
    \mathbf{r} = [r_0, r_1, \ldots, r_{K-1}] \in \mathbb{R}^{K}.
\end{equation}
To obtain differentiable ranks, we compute pairwise probabilities between all image pairs as:
\begin{equation}
    P_{ij} = \sigma\!\left(\frac{r_j - r_i}{\tau}\right),
\end{equation}
where $\sigma(\cdot)$ denotes the sigmoid function and $\tau$ is a temperature hyperparameter controlling the ranking sharpness.  
The predicted soft rank for each image is then defined as:
\begin{equation}
    \hat{R}_i = 1 + \sum_{j=0}^{K-1} P_{ij}.
\end{equation}
The ground-truth rank $\mathbf{R} = [1, 2, \ldots, K]$ reflects the monotonic decrease in color fidelity from the real image ($R_0=1$) to the most distorted synthetic sample ($R_{K-1}=K$).  
The training loss minimizes the mean squared error between predicted and ground-truth ranks:
\begin{equation}
    \mathcal{L} = \frac{1}{K}\sum_{i=0}^{K-1} (\hat{R}_i - R_i)^2.
\end{equation}
}

\fang{
This \textit{softrank loss} provides stable and differentiable supervision for learning perceptual color orderings, encouraging the model to assign higher scores to visually realistic images while penalizing over-saturated or distorted ones. 
It effectively bridges the continuous nature of color fidelity with ordinal supervision derived from the guidance-controlled image sequences in CFD.
}

\section{Color Fidelity Refinement}
\label{sec:realism_pipeline}
\fang{
We further introduce a training-free \textit{Color Fidelity Refinement} (CFR) pipeline that adjusts the guidance scale in diffusion-based T2I models spatially and temporally based on cross-modal attention. In our formulation, the attention map indicates the degree of \textit{color realism discrepancy} between the generated image and natural photographic appearance. Regions with higher attention responses are those whose color appearance deviates from the characteristics of natural photographic imagery.
}

\noindent\textbf{Attention from multimodal embeddings.}
\fang{
Let $\mathbf{F}^{v} = [\mathbf{f}_1^{v}, \ldots, \mathbf{f}_M^{v}] \in \mathbb{R}^{M \times d}$ 
and $\mathbf{F}^{t} = [\mathbf{f}_1^{t}, \ldots, \mathbf{f}_N^{t}] \in \mathbb{R}^{N \times d}$ 
denote the normalized visual and textual token embeddings extracted from the Qwen2-VL backbone. 
We compute the text-to-image attention matrix with a temperature parameter $\kappa > 0$ as
\begin{equation}
\mathbf{A} = \text{softmax}\!\left(\frac{\mathbf{F}^{t} (\mathbf{F}^{v})^{\top}}{\kappa}\right) \in \mathbb{R}^{N \times M},
\end{equation}
where $\mathbf{A}_{ij}$ measures the response between the $i$-th text token and the $j$-th visual token.
We then aggregate token-level responses across all text tokens to obtain a per-visual-token attention map:
\begin{equation}
\mathbf{a} = \frac{1}{N}\sum_{i=1}^{N} \mathbf{A}_{i:} \in \mathbb{R}^{M}.
\end{equation}
The one-dimensional attention sequence $\mathbf{a}$ is reshaped into a two-dimensional spatial map, linearly normalized to the range [0, 1] and upsampled as $\mathbf{a}'$ to match the latent resolution of the generation process, ensuring spatial alignment with the model’s latent representation for subsequent color fidelity refinement.
}

\noindent\textbf{Spatial--temporal guidance modulation.}
\fang{
Let $s_0$ be the base classifier-free guidance scale and $T$ the total number of denoising steps. 
CFR produces a spatially varying guidance field $s_t(u,v)$ that decays over time and is attenuated more strongly in highly attended regions:
\begin{equation}
s_t(u,v) \;=\;  s_0 \,\big[ 1 - \lambda\,\alpha(t)\,\mathbf{a}'(u,v) \big],
\end{equation}
where $\lambda \in [0,1]$ controls the modulation strength, 
and $\alpha(t) = 1 - \tfrac{t}{T}$ defines a temporal decay factor that gradually reduces the overall modulation intensity along the denoising trajectory. 
In this way, the effective guidance scale dynamically decreases in regions with high color–semantic discrepancy while remaining close to $s_0$ elsewhere, enabling color correction without compromising semantic consistency.
}

\noindent\textbf{Refined denoising update.}
\fang{
As discussed in Sec.~\ref{sec:cfg}, standard CFG~\cite{ho2022classifier} combines conditional and unconditional noise predictions using a fixed global guidance scale $s$ to control semantic adherence, as formulated in Eq.~\ref{eqa:cfg}. In our formulation, we extend this mechanism by introducing a spatially varying and temporally decaying guidance field $s_t(u,v)$ that adapts to local color–semantic discrepancies. 
At each denoising step $t$, the refined noise prediction is computed as:
\begin{equation}
\begin{aligned}
\hat{\epsilon}_{\theta}(u,v) 
&= \epsilon_{\theta}(\mathbf{z}_t)(u,v) \\
&\quad +\, s_t(u,v)\,\Big(\epsilon_{\theta}(\mathbf{z}_t,\mathbf{c})(u,v) 
      - \epsilon_{\theta}(\mathbf{z}_t)(u,v)\Big),
\end{aligned}
\end{equation}
and the latent is subsequently updated (omitting sampler-specific constants for brevity) as:
\begin{equation}
\mathbf{z}_{t-1} = \Phi\!\big(\mathbf{z}_t,\, \hat{\epsilon}_{\theta},\, t\big).
\end{equation}
}
\label{experiment}
\begin{figure*}
    \centering
    \includegraphics[width=1\linewidth]{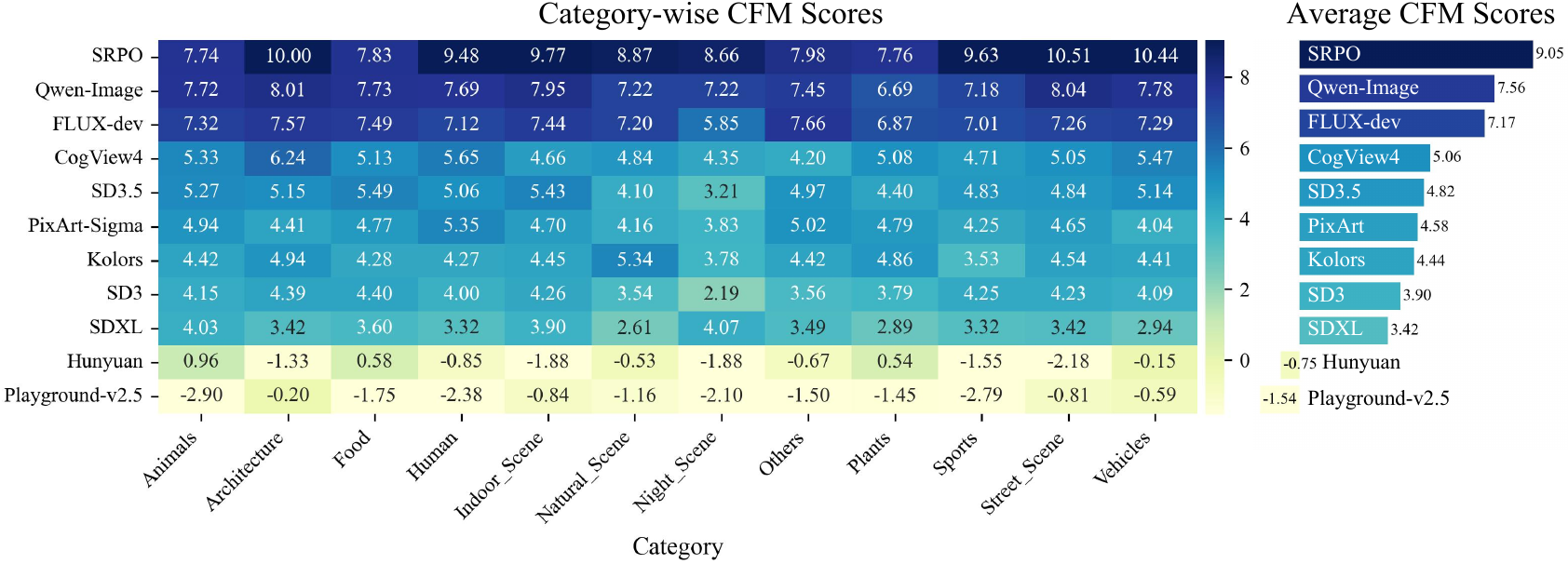}
    \caption{
\textbf{Benchmark results of color fidelity evaluation across different text-to-image models.}
Left: category-wise CFM scores across 12 categories.  Right: average CFM scores obtained by each model using our benchmark. }

    \label{fig:benchmark}
    \vspace{-8pt}
\end{figure*}

\noindent\textbf{Discussion.}
Unlike the standard CFG that applies a uniform scale across the entire image, our spatial–temporal guidance dynamically adapts to both location and timestep. This mechanism suppresses guidance in regions with high color–fidelity discrepancy (e.g., over-saturated) while preserving color realism elsewhere, effectively mitigating artifacts such as over-enhancement and contrast imbalance. As CFR relies solely on attention-based modulation without altering model parameters, it functions as a fully plug-and-play refinement module compatible with any diffusion-based T2I model. Further implementation details are provided in the supplementary material.

%% file: sec/5_experiment.tex
\section{Experiment}
\subsection{Implementation Details}
\noindent\textbf{Datasets.} 
We collect a total of 189,490 high-quality real-world photographs from publicly available datasets such as COCO and Open Images. 
All images are filtered using CLIPIQA~\cite{wang2023exploring} and the Qwen2.5-VL (72B) model~\cite{qwen2.5} to ensure authentic photographic quality, 
and resized such that the longer side does not exceed 1024 pixels. 
To construct the CFD-Training dataset containing 1.12M images, 
we employ multiple T2I models, including 
SDXL~\cite{podell2023sdxl}, SD3~\cite{esser2024sd3}, SD3.5, PixArt-$\Sigma$~\cite{chen2024pixart},
Kolors~\cite{kolors}, CogView4~\cite{ding2021cogview}, and Hunyuan-DiT~\cite{li2024hunyuan}. 
We further extend the benchmark by incorporating generations from 
Flux-dev~\cite{flux2024}, Qwen-Image~\cite{wu2025qwenimagetechnicalreport}, 
Playground-v2.5~\cite{li2024playground}, and SRPO~\cite{shen2025directly} 
to build the CFD-Test and CFD-Human datasets.

\noindent\textbf{Metrics.} 
We evaluate a wide range of T2I models using the proposed Color Fidelity Metric (CFM). 
For each model, 12,000 images are generated across 12 semantic categories, 
with all generations conditioned on an identical prompt set to ensure fair cross-model comparability. 
To verify the effectiveness of CFM, we conduct two experiments: 
1) measuring \textit{discrimination accuracy} on the CFD-Test dataset under two configurations, and 
2) computing the \textit{Spearman}, \textit{Pearson}, and \textit{Kendall} correlation coefficients 
between CFM predictions and human ratings on the CFD-Human dataset. 
Finally, for the overall evaluation of the proposed Color Fidelity Refinement (CFR) results, 
we report FID~\cite{heusel2017gans}, CLIPScore~\cite{hessel2021clipscore}, $\Delta$Sat. , and CFM, 
where $\Delta$Sat. denotes the absolute difference from the real-image saturation of 0.33.

\noindent\textbf{Other details.} 
All model training and image generation are performed on NVIDIA H20 GPUs. 
CFM is fine-tuned for 1 epoch with a batch size of 32, a learning rate of $2\times10^{-6}$, 
and a warm-up ratio of 0.05. 
Additional implementation details are provided in the supplementary material.

\subsection{Benchmark and Evaluation}
\label{sec:benchmark}
We employ the proposed CFM as an objective metric to assess the color realism of diverse T2I models.
The evaluation is conducted on the CFD, which provides controlled supervision of color fidelity under consistent textual prompts and balanced category distributions.
As shown in Fig.~\ref{fig:benchmark}, our benchmark reveals clear performance variation across different generation paradigms. The recent SRPO~\cite{shen2025directly} achieves the highest overall CFM score of 9.05, demonstrating its remarkable effectiveness in enhancing the realism of generated images. Models such as Qwen-Image~\cite{wu2025qwenimagetechnicalreport} and Flux-dev also demonstrate competitive performance, whereas models that emphasize aesthetic optimization (e.g., Playground-v2.5~\cite{li2024playground}) tend to produce reduced color fidelity. Additional visualization and cross-benchmark evaluations are provided in the supplementary material.

\subsection{Experimental Validation of Effectiveness}
\label{sec:validation}

\input{tab/accuracy}
We evaluate the effectiveness of the proposed Color Fidelity Metric (CFM) from two perspectives: its accuracy in selecting the image with higher color realism from each pair, and its alignment with human perceptual judgments regarding color authenticity. Further implementation details are provided in the supplementary material.

\noindent\textbf{Color fidelity discrimination accuracy.}
\fang{
We evaluate the effectiveness of CFM through pairwise discrimination experiments on the CFD-Test dataset under two settings: 1) CFD-SynPairs, consisting of synthetic image pairs from the same group, and 2) CFD-Real\&Syn, containing pairs of a real image and a randomly sampled synthetic counterpart.
As shown in Tab.~\ref{tab:accuracy}, CFM achieves over 80\% accuracy in both settings, clearly demonstrating its strong ability to distinguish color authenticity across real and synthetic domains.
In contrast, existing T2I aesthetic metrics show lower accuracy due to their bias toward vivid, high-contrast imagery~(Sec.~\ref{sec:intro}), often undervaluing naturally balanced images with realistic color tones.
Traditional image quality assessment methods~\cite{ke2021musiq,yang2022maniqa,wang2023exploring,zhang2023blind} perform near-randomly on synthetic pairs and remain considerably inferior on real–synthetic comparisons, as they are primarily designed for real-image degradations and fail to generalize to generative domains.
}

\noindent\textbf{Human consistency correlation.}
\fang{
To further evaluate how well CFM align with human perception of color realism, we compute Spearman, Pearson, and Kendall correlation coefficients between predicted scores and human ratings on the CFD-Human dataset, as summarized in Tab.~\ref{tab:spearman}.
Our proposed CFM achieves the highest correlation across all, surpassing existing aesthetic-based metrics by a clear margin.
This indicates that CFM better captures the perceptual consistency underlying human evaluation of color authenticity.
In contrast, other metrics exhibit weaker correlation, as they tend to emphasize global semantic relevance or aesthetic attractiveness rather than faithful color reproduction.
}

\fang{
These results highlight that CFM not only discriminates color realism effectively but also maintains strong alignment with human perceptual judgments, validating its reliability as an objective measure for evaluating photorealistic quality in text-to-image generation.
}
\begin{figure}
    \centering
    \includegraphics[width=1\linewidth]{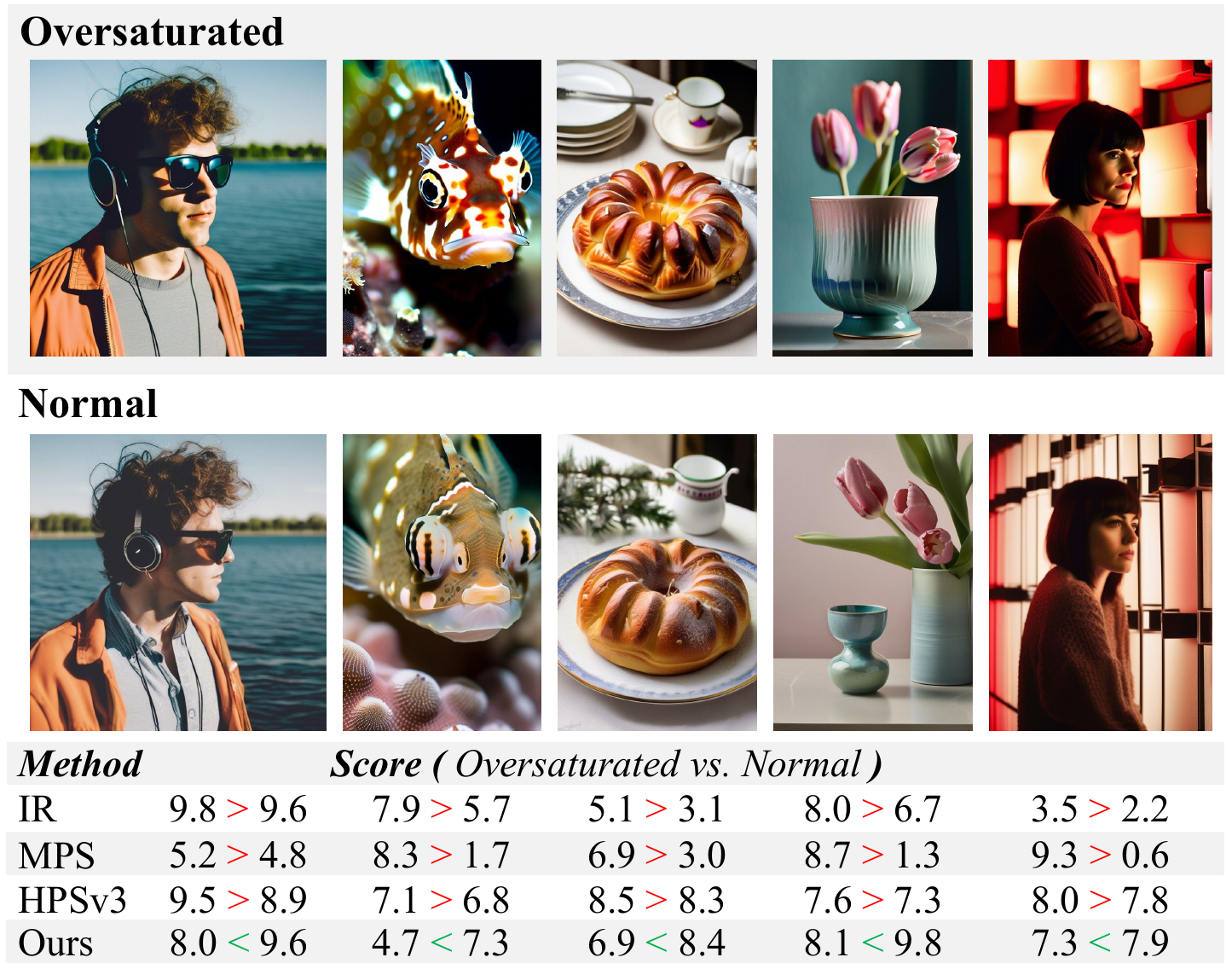}
    \caption{
    \textbf{Examples of evaluation bias between existing metrics and ours.} 
    Existing metrics perform less favorably due to their training bias toward vivid, high-contrast images, whereas our proposed CFM provides a more accurate assessment of color realism, assigning higher scores to images with naturally balanced and authentic colors.}
    \label{fig:metric_vis}
    \vspace{-6pt}
\end{figure}
\input{tab/spearman}

\subsection{Results of CFR}
\fang{
The proposed Color Fidelity Refinement (CFR) pipeline for color fidelity enhancement can be seamlessly integrated into any diffusion-based generative model employing the standard classifier-free guidance mechanism.
As shown in Tab.~\ref{tab:agr_results}, we apply the CFM-based CFR pipeline to SD3.5~\cite{esser2024scaling}, PixArt-$\Sigma$~\cite{chen2024pixart}, and Hunyuan-DiT~\cite{li2024hunyuan}, and report the quantitative results before and after enhancement in terms of FID~\cite{heusel2017gans}, CLIPScore~\cite{hessel2021clipscore}, saturation difference ($\Delta$Sat.) and CFM.
Compared with the original outputs, models equipped with the CFM-driven CFR produce images with more natural and visually realistic color appearances.
As shown in Tab.~\ref{tab:agr_results},  saturation values decrease by 0.08–0.11 and CFM scores increase by 1.3–2.0 points across all three models, while maintaining comparable image quality (FID) and semantic alignment (CLIPScore).
}
\begin{figure}
    \centering
    \includegraphics[width=1\linewidth]{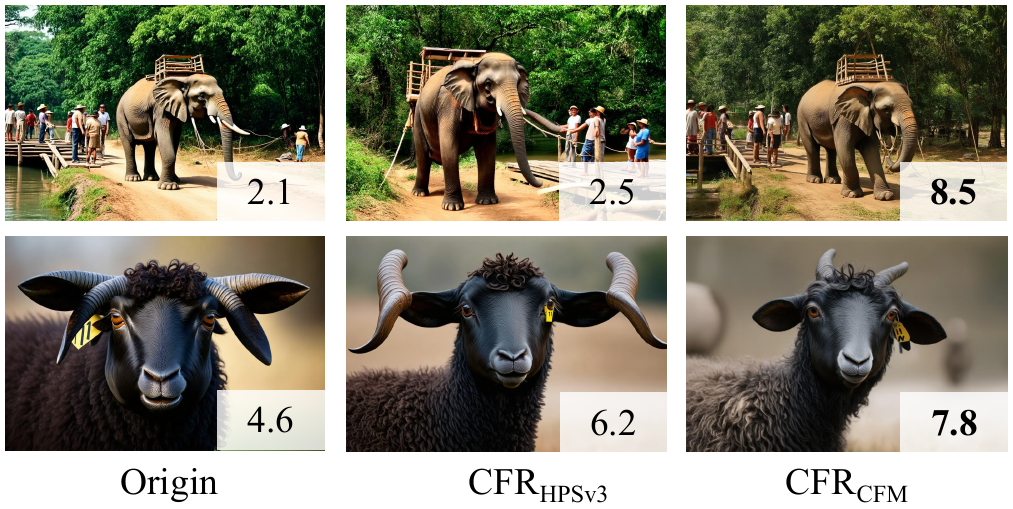}
    \caption{\textbf{Qualitative examples of different CFR variants.} `Origin' denotes the original image before applying CFR, while `CFR' represents the results enhanced by the corresponding model-based CFR. The bottom-right value in each image indicates its CFM score, where a higher score reflects better color fidelity.}
    \label{fig:agr_vis}
    \vspace{-5pt}
\end{figure}
\input{tab/agr}

\fang{
To further validate the effectiveness of the attention derived from the color fidelity–oriented CFM, we also conduct experiments with an HPSv3-based CFR variant.
As illustrated in Tab.~\ref{tab:agr_results}, although HPSv3-based CFR provides moderate improvements in color appearance, its enhancement effect is notably weaker than that of the CFM-based approach. In addition, we provide qualitative examples of both CFR variants in Fig.~\ref{fig:agr_vis}.
As shown, CFR$_{\text{CFM}}$ effectively reduces oversaturation and excessive contrast, producing images with more balanced and natural colors, accompanied by higher CFM scores.
In contrast, CFR$_{\text{HPSv3}}$ still exhibits slight oversaturation artifacts, resulting in lower CFM scores and less color fidelity. More results are provided in the supplementary material.
}

\subsection{Ablation Study}
\noindent\textbf{CFM: Softrank loss vs. pairwise loss.}
To evaluate the advantage of the proposed differentiable softrank loss, we replace it with a binary pairwise ranking objective that minimizes $\log(1+\exp(-(r_i - r_j)))$ across image pairs.
Both variants are trained on the same CFD-Training set using identical backbones, learning rates, and epochs.
As summarized in Tab.~\ref{tab:ablation_cfs} (rows~1 and~3), the proposed softrank loss (`Ours') achieves considerably higher discrimination accuracy on both CFD-SynPairs and CFD-Real\&Syn, and exhibits stronger correlation with human ratings on CFD-Human. 
This result demonstrates that the continuous, differentiable ranking formulation is essential for learning robust color-realism orderings.

\input{tab/ab_cfm}

\noindent\textbf{CFM: Role of textual conditioning.}
We further ablate the multimodal structure of CFM by removing the textual branch and training a visual-only variant that uses identical visual embeddings and MLP heads but omits all text tokens.
As shown in Tab.~\ref{tab:ablation_cfs}, rows~2 and~3, removing textual conditioning leads to a notable performance drop (–6.5\% on CFD-Test SynPairs accuracy and –6 Spearman correlation). 
This suggests that textual context provides crucial cues for interpreting expected color distributions, enabling the model to distinguish between perceptually appropriate and distorted tones under different semantic conditions.
These findings validate the importance of text–image joint representation in modeling color–semantic consistency.

\begin{figure}
    \centering
    \includegraphics[width=1\linewidth]{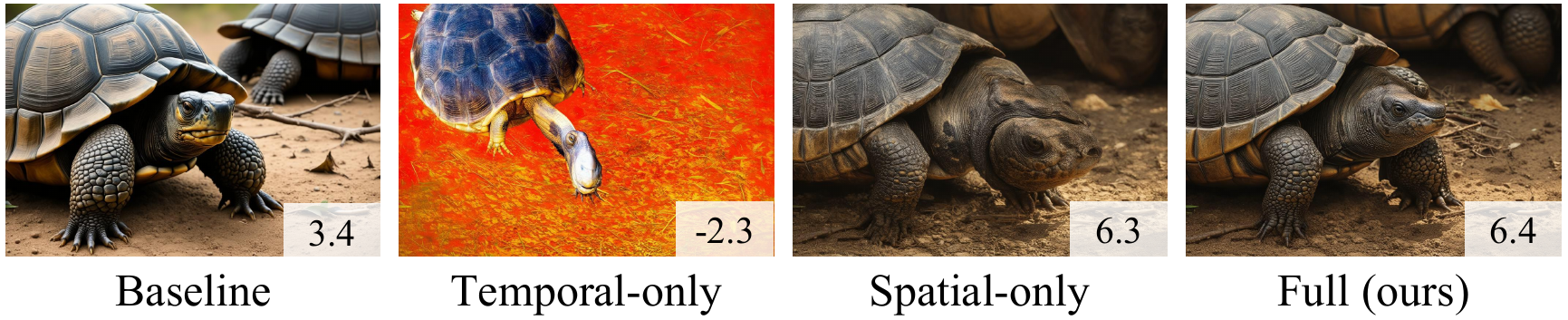}
    \caption{\textbf{Qualitative comparison of different CFR variants on SD3.5.} The temporal-only variant causes semantic inconsistency and severely distorted colors, while the spatial-only variant mitigates over-saturation but yields suboptimal detail preservation. The full spatial–temporal CFR achieves balanced and natural color rendering with the highest CFM scores.}
    \label{fig:ab_agr}
    \vspace{-7pt}
\end{figure}

\noindent\textbf{CFR: Spatial–temporal modulation.}
\fang{
Finally, we investigate the contribution of the spatial and temporal components in the proposed CFR mechanism in Sec.~\ref{sec:realism_pipeline}. 
We compare four variants on SD3.5~\cite{esser2024sd3}: 1) the baseline with a fixed global guidance scale, 2) temporal-only modulation that decays uniformly over timesteps, 3) spatial-only modulation based on attention but without temporal decay, and 4) the full spatial–temporal CFR.
As shown in Tab.~\ref{tab:ablation_agr_components} and Fig.~\ref{fig:ab_agr}, the temporal-only variant fails to provide stable enhancement and even degrades performance, leading to higher FID (+4.7), lower CLIPScore (–2.3), and negative CFM values, indicating that time-decay modulation alone weakens the conditional guidance strength of CFG, thereby disrupting semantic consistency without effectively improving color fidelity.
In contrast, the spatial-only variant improves both FID and CFM, demonstrating that attention-based spatial modulation plays a central role in correcting over-saturated regions.
The full CFR achieves the best results, showing that temporal decay stabilizes and complements spatial attention to consistently enhance color realism without compromising semantic fidelity.
}
\input{tab/ab_agr}


%% file: tab/accuracy.tex
\begin{table}[t]
  \centering
   \caption{
  Color fidelity prediction accuracy (\%) on CFD-Test.}
  \vspace{-5pt}
  \setlength\tabcolsep{2mm}{
    \scalebox{0.7}{
  \begin{tabular}{l|cc}
    \toprule
    \textbf{Model} & \textbf{CFD-SynPairs} & \textbf{CFD-Real\&Syn} \\
    \midrule
    \multicolumn{3}{c}{\textit{Traditional image quality measures}} \\
    \midrule
    MUSIQ~\cite{ke2021musiq} & 53.4 & 21.5 \\
    MANIQA~\cite{yang2022maniqa} & 54.5 & 24.0 \\
    CLIPIQA~\cite{wang2023exploring} & 44.7 & 22.7 \\
    LIQE~\cite{zhang2023blind} & 50.1 & 23.6 \\
    \midrule
    \multicolumn{3}{c}{\textit{Generative aesthetic scoring methods}} \\
    \midrule
    ImageReward~\cite{xu2023imagereward} & 44.3 & 42.7 \\
    PickScore~\cite{kirstain2023pick} & 51.4 & 48.5 \\  
    MPS~\cite{zhang2024learning} & 44.4 & 46.2 \\
    HPSv3~\cite{ma2025hpsv3} & 57.5 & 58.3 \\
    \midrule
    \textbf{Ours} & \textbf{83.6} & \textbf{80.1} \\
    \bottomrule
  \end{tabular}
  }}
  \label{tab:accuracy}
  \vspace{-10pt}
\end{table}

%% file: tab/spearman.tex
\begin{table}[t]
\centering
\caption{Correlation of CFM scores on the CFD-Human.}
\vspace{-5pt}
\small
\scalebox{0.8}{
\begin{tabular}{l|ccc}
\toprule
\textbf{Metric} &\textbf{Spearman} ($\rho$)$\uparrow$ & \textbf{Pearson} ($r$)$\uparrow$  & \textbf{Kendall} ($\tau$)$\uparrow$   \\
\midrule
ImageReward~\cite{xu2023imagereward} & 62.8 & 63.5 & 49.2 \\
PickScore~\cite{kirstain2023pick}    & 71.2 & 72.1 & 56.5 \\
MPS~\cite{zhang2024learning}         & 66.7 & 78.5 & 52.8 \\
HPSv3~\cite{ma2025hpsv3}             & 74.4 & 76.0 & 62.8 \\
\midrule
\textbf{Ours} &\textbf{84.9} &\textbf{85.4}  &  \textbf{71.4}   \\  
\bottomrule
\end{tabular}}
\vspace{-2mm}
\label{tab:spearman}
\vspace{-4pt}
\end{table}

%% file: tab/agr.tex
\begin{table}[t]
  \centering
    \caption{Quantitative results of CFR.}
    \vspace{-7pt}
  \setlength{\tabcolsep}{3mm}{
  \scalebox{0.7}{
  \begin{tabular}{l|l|cccc}
    \toprule
    \textbf{Model} & \textbf{Setting} & \textbf{FID}$\downarrow$ & \textbf{CLIPScore}$\uparrow$ & $\boldsymbol{\Delta}$\textbf{Sat.}$\downarrow$ & \textbf{CFM}$\uparrow$ \\
    \midrule
    \multirow{3}{*}{SD3.5~\cite{esser2024scaling}} 
      & $-$ &  13.3  &  28.2  &  0.15  &  4.9  \\
      & CFR$_{\text{HPSv3}}$ &  13.2  &  28.1  &  0.11  &  5.6  \\
      & \textbf{CFR$_{\text{CFM}}$} &  \textbf{13.1}  &  \textbf{28.2}  &  \textbf{0.07}  &  \textbf{6.9}  \\
    \midrule
    \multirow{3}{*}{PixArt-$\Sigma$~\cite{chen2024pixart}} 
      & $-$ &  16.5  &  27.2  &  0.09  &  4.4  \\
      & CFR$_{\text{HPSv3}}$ &  16.4  &  27.4  &  0.05  &  5.1  \\
      & \textbf{CFR$_{\text{CFM}}$} &  \textbf{16.4}  &  \textbf{27.5}  &  \textbf{0.02}  &  \textbf{6.4}  \\
    \midrule
    \multirow{3}{*}{Hunyuan~\cite{li2024hunyuan}} 
      & $-$ &  22.1  &  27.5  &  0.14  &  0.8  \\
      & CFR$_{\text{HPSv3}}$ &  22.0  &  27.5  &  0.06  &  1.7  \\
      & \textbf{CFR$_{\text{CFM}}$} &  \textbf{19.9}  &  \textbf{27.5}  &  \textbf{0.03}  &  \textbf{2.1}  \\
    \bottomrule
  \end{tabular}
  }}
  \label{tab:agr_results}
  \vspace{-5pt}
\end{table}

%% file: tab/ab_cfm.tex
\begin{table}[t]
\centering
\caption{Ablation on CFM components.}
\vspace{-7pt}
\small
\setlength{\tabcolsep}{4.2pt}
\scalebox{0.8}{
\begin{tabular}{l|cccc}
\toprule
\multirow{2}{*}{\textbf{Variant}} & \multicolumn{2}{c}{\textbf{CFD-Test Acc. (\%)}} & \multicolumn{2}{c}{\textbf{CFD-Human Corr.}} \\
\cmidrule(lr){2-3}\cmidrule(lr){4-5}
& \textbf{SynPairs} & \textbf{Real\&Syn} & \textbf{Spearman $\uparrow$} & \textbf{Kendall} $\uparrow$ \\
\midrule
Pairwise loss & 76.2 & 74.8 & 78.3 & 66.1 \\
Visual-only & 77.1 & 74.3 & 78.9 & 67.0 \\
\textbf{Ours} & \textbf{83.6} & \textbf{80.1} & \textbf{84.9} & \textbf{71.4} \\
\bottomrule
\end{tabular}}
\label{tab:ablation_cfs}
\end{table}

%% file: tab/ab_agr.tex
\begin{table}[t]
\centering
\caption{Ablation on CFR: Spatial–temporal modulation.}
\vspace{-7pt}
\small
\setlength{\tabcolsep}{3.8pt}
\scalebox{0.8}{
\begin{tabular}{l|cccc}
\toprule
\textbf{Setting} & \textbf{FID}$\downarrow$ & \textbf{CLIPScore}$\uparrow$ & \textbf{$\Delta$Sat.}$\downarrow$ & \textbf{CFM}$\uparrow$ \\
\midrule
Baseline ($s_0$) & 13.3 & 28.2 & 0.15 & 4.9 \\
Temporal-only    & 18.0 & 25.9 & 0.18 & -1.3 \\
Spatial-only     & 13.2 & 28.2 & 0.12 & 6.8 \\
Full (ours)      & \textbf{13.0} & \textbf{28.2} & \textbf{0.07} & \textbf{6.9} \\
\bottomrule
\end{tabular}}

\label{tab:ablation_agr_components}
\end{table}

%% file: sec/6_conclusion.tex
\section{Conclusion}
\label{sec:conclusion}
In this work, we tackle the overlooked challenge of \textit{color fidelity} in realistic-style T2I generation. 
While modern T2I models achieve strong semantic alignment, their outputs often appear unrealistically vivid. 
We introduce the \textit{Color Fidelity Dataset (CFD)}, the \textit{Color Fidelity Metric (CFM)}, and the \textit{Color Fidelity Refinement (CFR)}, a comprehensive framework for evaluating and enhancing color realism. 
CFM aligns closely with human perception, achieving over 80\% discrimination accuracy, while CFR effectively improves color fidelity without compromising image quality and semantic consistency.
Together, they provide the first unified solution for quantifying and improving perceptual color authenticity in realistic-style image generation.

%% file: sec/ack.tex
\section*{Acknowledgements}
This work was supported by the National Natural Science Foundation of China (Grant No. 62372133, 62125201 and U24B20174).

%% file: sec/X_suppl.tex
\clearpage
\setcounter{page}{1}
\maketitlesupplementary

\section*{A. Dataset Details}
\label{sec:supp_dataset}
\fang{
To support objective evaluation of color fidelity in realistic-style text-to-image (T2I) generation, 
we construct the large-scale \textbf{Color Fidelity Dataset (CFD)}, which provides high-quality real-world photographs, 
their corresponding textual captions, and progressively distorted synthetic counterparts generated under different guidance scales. 
This section provides detailed statistics and construction protocols for the dataset, 
including category distribution, model composition, and benchmark splits. 
}

As summarized in Tab.~\ref{tab:cfd_category_distribution}, the \textbf{Color Fidelity Dataset (CFD)} comprises 189,490 high-quality real-world photographs evenly distributed across 12 semantic categories, ensuring comprehensive coverage of diverse color appearances encountered in everyday photography. These real images are carefully curated from open-source photographic collections with rigorous filtering to remove synthetic, over-processed, or low-resolution samples. Each image serves as a perceptual anchor for evaluating the color fidelity of its synthetic counterparts.

\begin{figure*}
    \centering
    \includegraphics[width=0.9\linewidth]{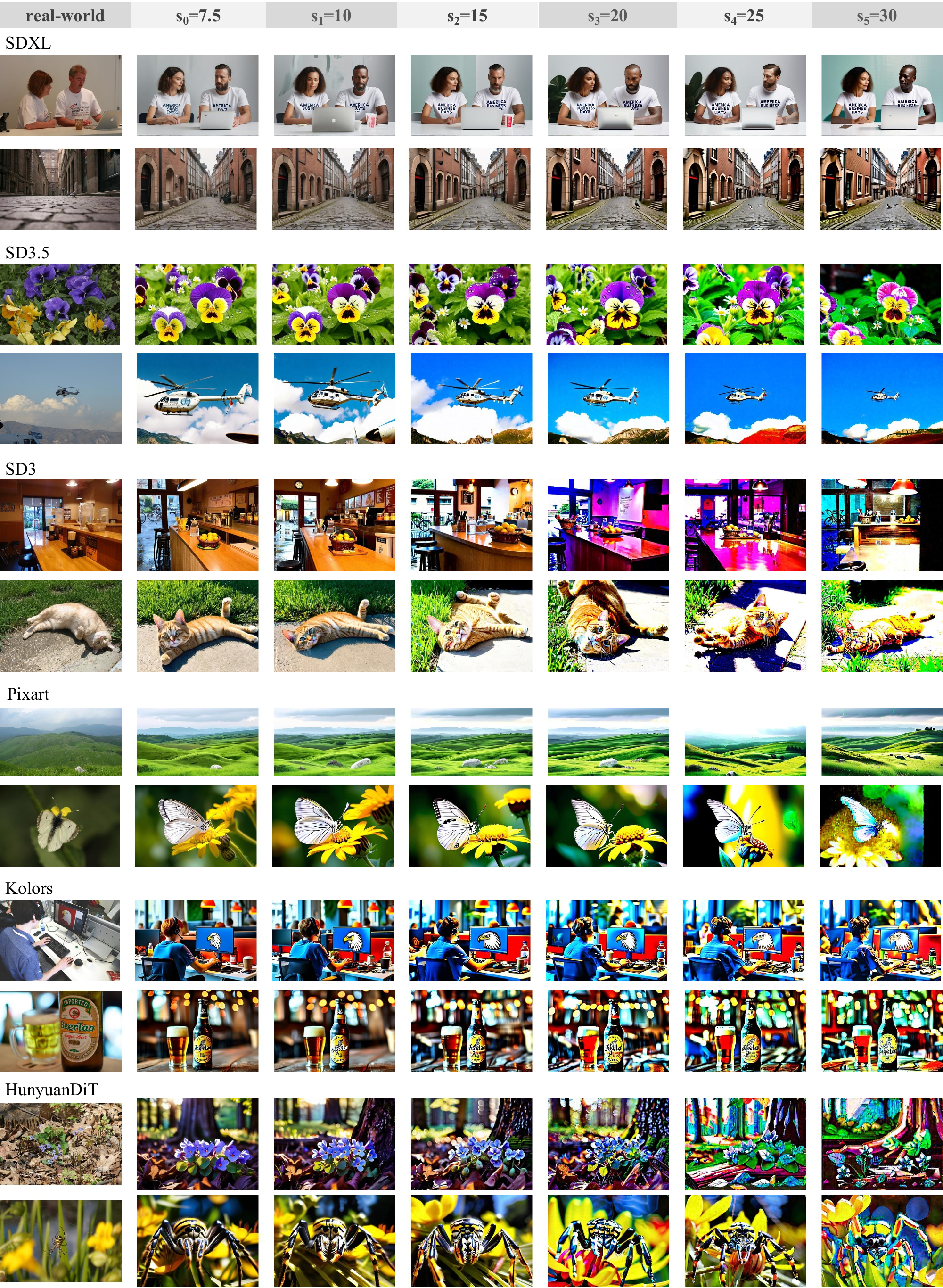}
    \caption{Visualization examples of CFD.}
    \label{fig:dataset_vis}
\end{figure*}

Tab.~\ref{tab:cfd_models_dualcol} further details the construction of the training and testing splits. Each real image is associated with six progressively distorted synthetic images generated under different classifier-free guidance (CFG) scales ($s \in {7.5, 10, 15, 20, 25, 30}$), enabling ordered supervision of color realism levels. The \textbf{CFD-Train} subset contains 160,000 real photographs and over 1.1M synthetic images produced by seven representative text-to-image models, covering a balanced spectrum of diffusion-based architectures and training paradigms. The \textbf{CFD-Test} subset extends the model diversity to eleven generation systems, incorporating both commercial and open-source models to facilitate fair benchmarking across different generation pipelines. We also provide several visualization examples of CFD in Fig.~\ref{fig:dataset_vis} to illustrate the dataset composition and the progressive nature of the generated distortions.

In Sec.~\ref{sec:validation}, CFD-SynPairs is formed by randomly sampling pairs of synthetic images within the same group that correspond to \textit{adjacent} guidance scales (e.g., $s=10$ vs. $s=15$), which enables fine-grained assessment of color fidelity ranking consistency.
In contrast, CFD-Real\&Syn pairs each real image with its synthetic counterpart generated under the \textit{lowest} guidance scale ($s=7.5$), thus directly evaluating absolute color fidelity between real-world photographs and their minimally guided generations.
Both subsets contain 5{,}000 image pairs and are used for model discrimination accuracy analysis.

Additionally, in Sec.~\ref{sec:benchmark}, the \textbf{CFS Benchmark} for large-scale evaluation of color realism across generation models is constructed by sampling 1{,}000 real image–caption pairs from each of the 12 semantic categories within CFD-Test, resulting in a total of 12{,}000 samples.
This benchmark provides a balanced and category-diverse testbed for standardized comparison of fidelity scores and generation quality across different T2I systems.
\begin{figure}
    \centering
    \includegraphics[width=0.9\linewidth]{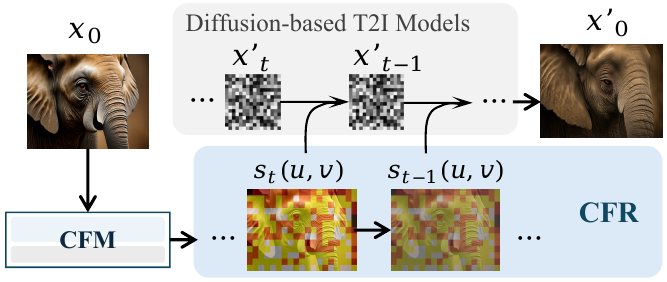}
    \caption{Framework of CFR.}
    \label{fig:cfr_pipeline}
\end{figure}

\input{tab/supp_train_test}

\begin{figure*}
    \centering
    \includegraphics[width=1\linewidth]{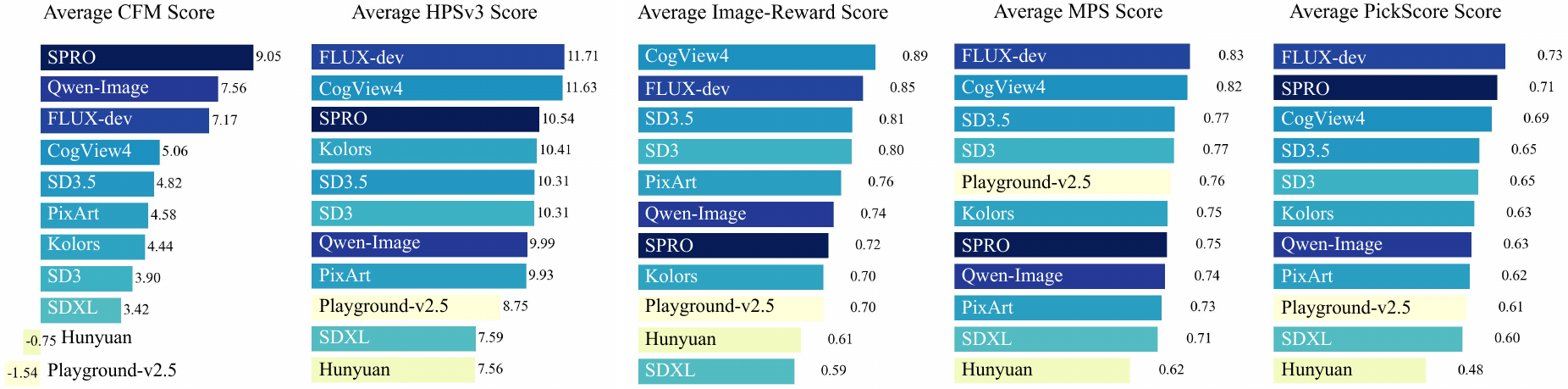}
    \caption{Cross-benchmark compairision.}
    \label{fig:benchmark_rank}
\end{figure*}

\input{tab/supp_realdata}

\section*{B. Benchmark Comparision and Visualization}
\begin{figure*}
    \centering
    \includegraphics[width=0.82\linewidth]{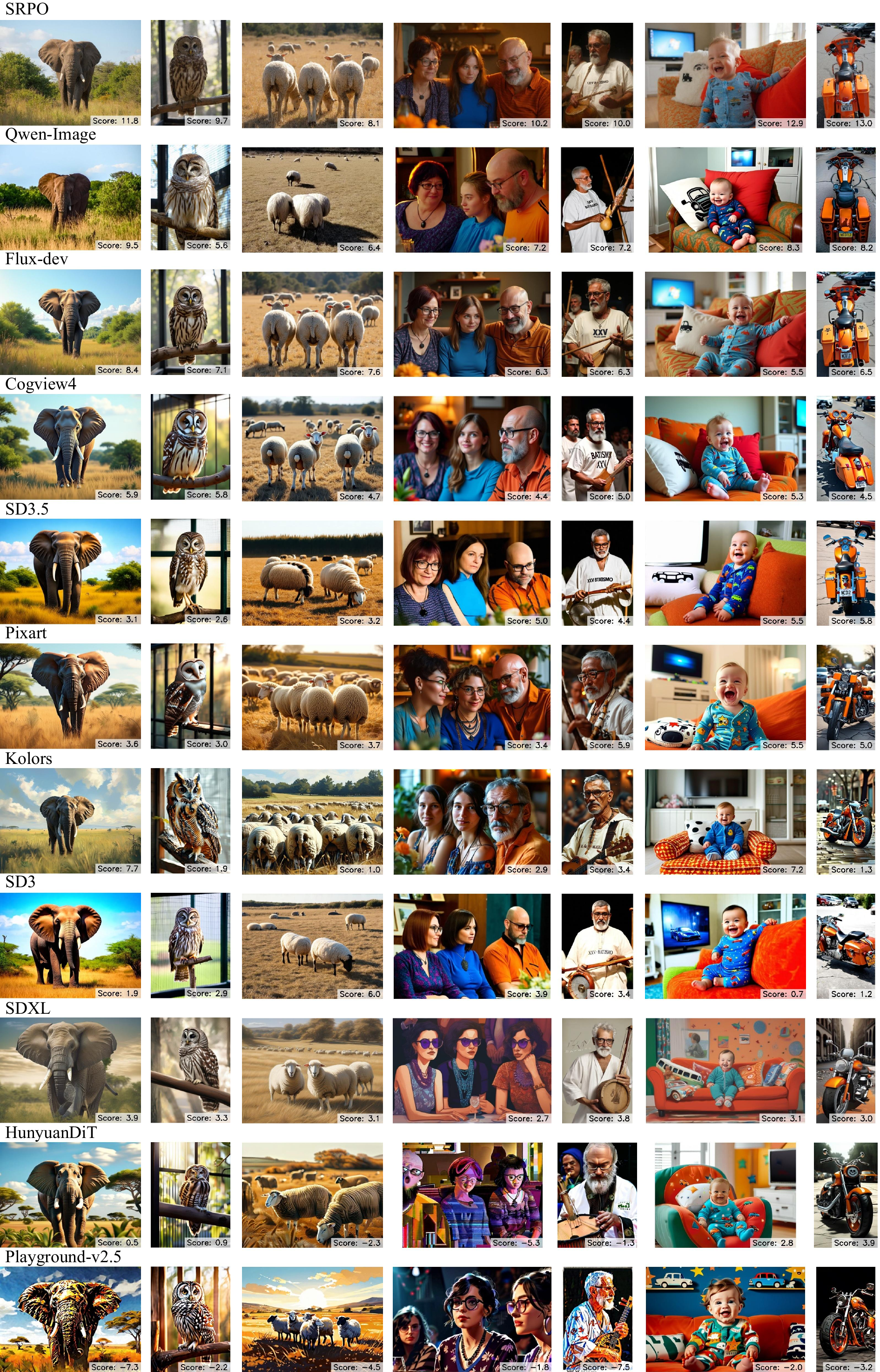}
    \caption{Visualization examples of CFM scores on realistic-style generations produced by different T2I models. The CFM score for each image is shown in the lower-right corner.}
    \label{fig:benchmark_compair}
\end{figure*}

\begin{figure*}
    \centering
    \includegraphics[width=0.82\linewidth]{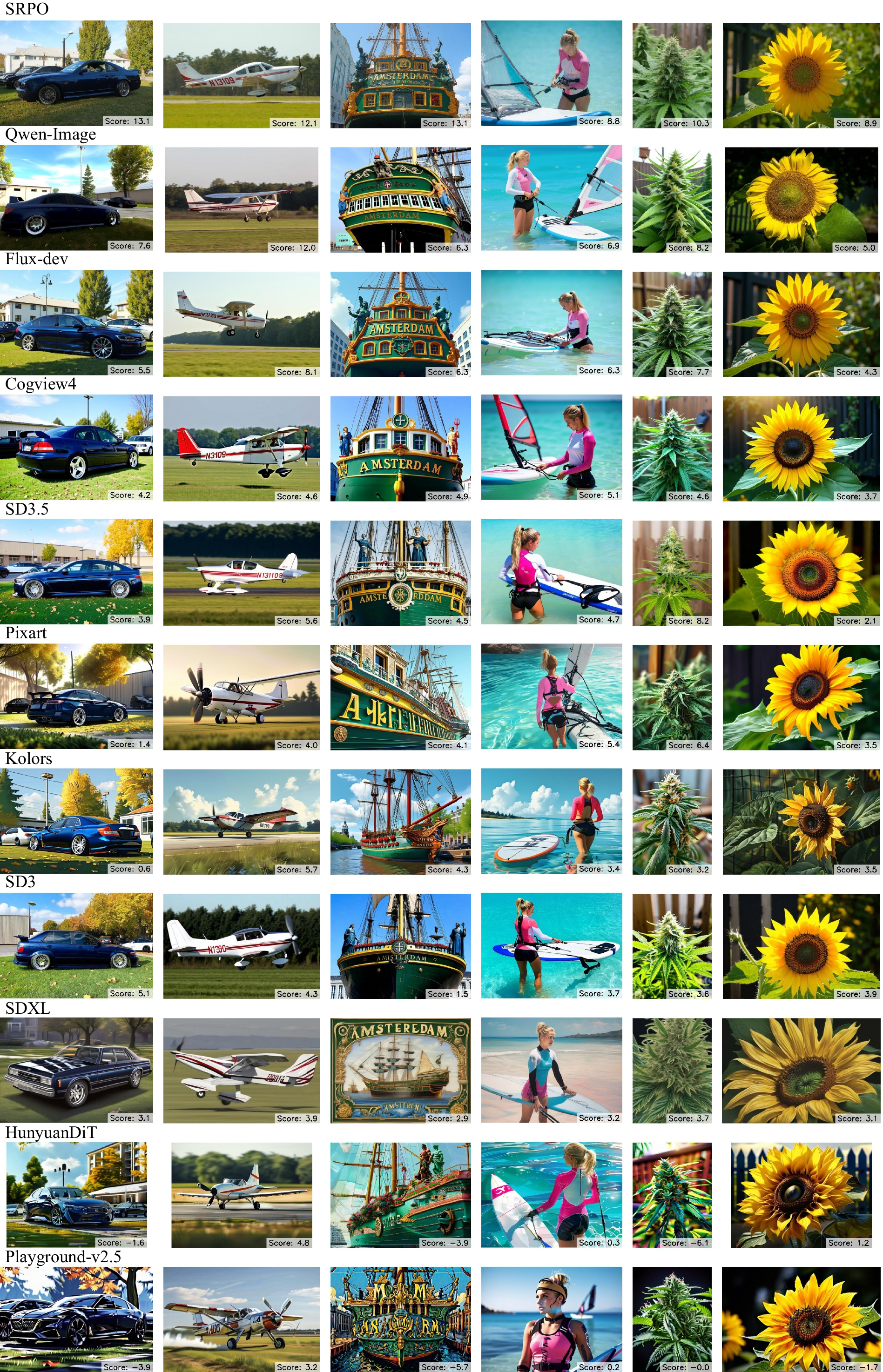}
    \caption{Visualization examples of CFM scores on realistic-style generations produced by different T2I models. The CFM score for each image is shown in the lower-right corner.}
    \label{fig:benchmark_compair2}
\end{figure*}

In Fig.~\ref{fig:benchmark_compair} and Fig.~\ref{fig:benchmark_compair2}, we present visualization examples of realistic-style generations produced by different T2I models, together with the corresponding scores assigned by our metric. Fig.~\ref{fig:benchmark_rank} further reports the ranking results of various models as evaluated by CFM and by existing metrics. It can be observed that our CFM provides substantially more accurate assessments of color fidelity, consistently reflecting the degree to which synthesized images preserve natural color appearance. In contrast, aesthetic-based or semantics-oriented metrics exhibit rankings that deviate significantly from the color realism of the generated images, as they primarily focus on visual appeal or text–image alignment rather than faithfully capturing color authenticity. These discrepancies highlight the necessity of a dedicated color-fidelity metric and demonstrate the reliability of our method.

\section*{C. Details of CFR}
To provide a clearer explanation of the Color Fidelity Refinement (CFR) mechanism introduced in Sec.~\ref{sec:realism_pipeline}, we illustrate the full refinement pipeline in Fig.~\ref{fig:cfr_pipeline}. CFR leverages the cross-modal attention extracted by the Color Fidelity Metric (CFM) to identify spatial regions that exhibit deviations from natural photographic color characteristics, and uses this information to modulate the guidance behavior of diffusion models.

Specifically, CFM computes text-to-image attention from the multimodal embeddings produced by the Qwen2-VL encoder. This attention distribution reflects the degree of \emph{color realism discrepancy} at each spatial location. After normalization and upsampling, the attention map serves as a spatial mask that adjusts the classifier-free guidance strength during the sampling process: areas with higher discrepancy receive stronger attenuation, while regions with naturally aligned color statistics preserve a guidance level closer to the original value. In addition, a temporal decay factor gradually reduces the modulation amplitude along the denoising trajectory, ensuring that the refinement does not disrupt semantic structure in later sampling steps.

At each denoising step, the diffusion model recomputes the noise prediction using the spatially and temporally varying guidance field, enabling fine-grained correction of color distortions without modifying model parameters. As a fully training-free method, CFR is compatible with any diffusion-based T2I model employing classifier-free guidance.

\begin{figure*}
    \centering
    \includegraphics[width=1\linewidth]{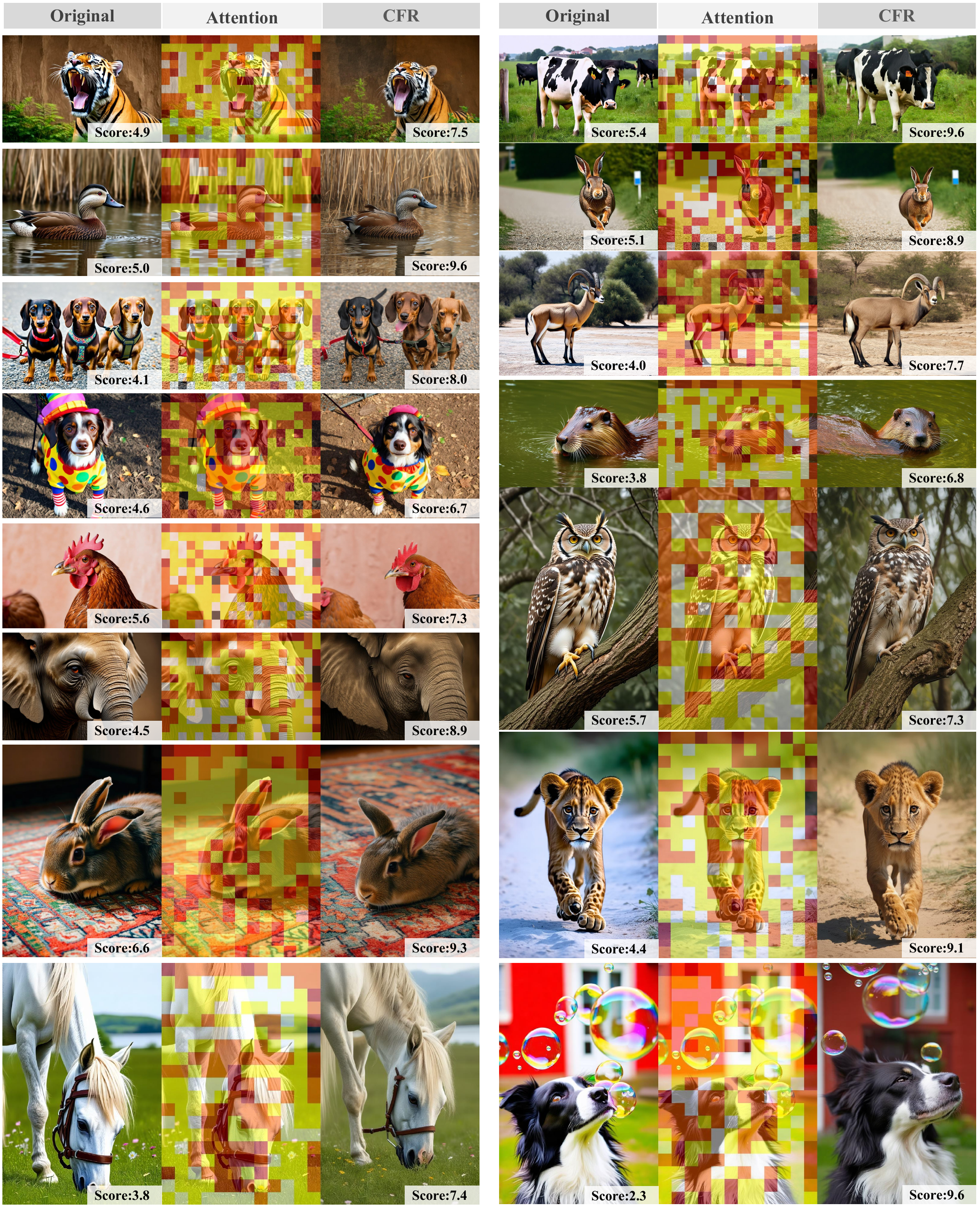}
    \caption{Visualization examples of CFR. The CFM score for each image is shown in the lower-right corner.}
    \label{fig:cfr_vis}
\end{figure*}

Furthermore, we visualize additional examples of CFR in Fig.~\ref{fig:cfr_vis}. As shown, CFR substantially improves the perceptual realism of generated images by mitigating oversaturation and restoring natural color appearance. Correspondingly, the CFM scores also display consistent increases, demonstrating the effectiveness of CFR in enhancing generative color fidelity.

\section*{D. Details of Implementation}
When training the CFM, we preserve the aspect ratio of each input image and resize it such that the longer edge is fixed to 448 pixels. We set the temperature hyperparameter of the softrank loss to $\tau = 0.1$. For the CFR pipeline, we use a temperature parameter of $\kappa = 10$ when computing the text-to-image attention matrix.

\section*{Limitations}
In this work, we exploit the varying distortion effects produced by different T2I models under multiple classifier-free guidance (CFG) scales to diversify color fidelity variations as much as possible. 
However, color distortions simulated solely through CFG adjustments remain limited in scope and may not fully represent the wide range of color deviations in real-world generative outputs. 
In future work, we plan to explore more comprehensive and controllable distortion mechanisms tailored for T2I models to achieve a more holistic evaluation of color fidelity.

%% file: tab/supp_train_test.tex
\begin{table*}[t]
\centering
\caption{Summary of CFD-Train and CFD-Test.}
\scalebox{0.82}{
\begin{tabular}{lccc}
\toprule
\textbf{Subset} & \textbf{Real Images} & \textbf{Total Images} & \textbf{Models Used} \\
\midrule
CFD-Train & 160{,}000 & 1{,}120{,}000 &
\begin{tabular}[c]{@{}l@{}}
SDXL~\cite{podell2023sdxl}, SD3~\cite{esser2024sd3}, SD3.5, PixArt-Sigma~\cite{chen2024pixart},
Kolors~\cite{kolors}, CogView4~\cite{ding2021cogview}, Hunyuan-DiT~\cite{li2024hunyuan}
\end{tabular} \\[4pt]
\midrule
CFD-Test & 29{,}490 & 206{,}430 &
\begin{tabular}[c]{@{}l@{}}
SDXL~\cite{podell2023sdxl}, SD3~\cite{esser2024sd3}, SD3.5, PixArt-Sigma~\cite{chen2024pixart}, Kolors~\cite{kolors},
CogView4~\cite{ding2021cogview}, Hunyuan-DiT~\cite{li2024hunyuan},\\ Flux-dev~\cite{flux2024}, 
Qwen-Image~\cite{wu2025qwenimagetechnicalreport}, Playground-v2.5~\cite{li2024playground}, SRPO~\cite{shen2025directly}
\end{tabular} \\
\bottomrule
\end{tabular}}
\label{tab:cfd_models_dualcol}
\end{table*}

%% file: tab/supp_realdata.tex
\begin{table}[t]
\centering
\caption{Category distribution of the Color Fidelity Dataset (CFD).}
\vspace{2mm}
\scalebox{0.9}{
\begin{tabular}{lcc}
\toprule
\textbf{Category} & \textbf{Real Images} & \textbf{Percentage (\%)} \\
\midrule
Human & 21{,}422 & 11.3 \\
Animals & 17{,}105 & 9.0 \\
Plants & 16{,}158 & 8.5 \\
Food & 17{,}474 & 9.2 \\
Vehicles & 13{,}631 & 7.2 \\
Sports & 9{,}963 & 5.3 \\
Architecture & 16{,}526 & 8.7 \\
Natural Scene & 15{,}947 & 8.4 \\
Street Scene & 17{,}211 & 9.1 \\
Indoor Scene & 17{,}895 & 9.4 \\
Night Scene & 12{,}369 & 6.5 \\
Others & 13{,}789 & 7.3 \\
\midrule
\textbf{Total} & \textbf{189{,}490} & \textbf{100.0} \\
\bottomrule
\end{tabular}}
\label{tab:cfd_category_distribution}
\end{table}

%% file: main.bib
@String(CVPR= {IEEE Conf. Comput. Vis. Pattern Recog.})

@String(ICCV= {Int. Conf. Comput. Vis.})

@String(ECCV= {Eur. Conf. Comput. Vis.})

@String(NIPS= {Adv. Neural Inform. Process. Syst.})

@String(TVCG  = {IEEE Trans. Vis. Comput. Graph.})

@String(ICLR = {Int. Conf. Learn. Represent.})

@String(AAAI = {AAAI})

@String(CVPR  = {CVPR})

@String(ICCV  = {ICCV})

@String(ECCV  = {ECCV})

@String(NIPS  = {NeurIPS})

@String(TVCG  = {IEEE TVCG})

@String(ICLR  = {ICLR})

@inproceedings{salimans2016improved,
  title={Improved techniques for training gans},
  author={Salimans, Tim and Goodfellow, Ian and Zaremba, Wojciech and others},
  booktitle={NIPS},
  year={2016}
}

@article{heusel2017gans,
  title={Gans trained by a two time-scale update rule converge to a local nash equilibrium},
  author={Heusel, Martin and Ramsauer, Hubert and Unterthiner, Thomas and Nessler, Bernhard and Hochreiter, Sepp},
  journal={NIPS},
  volume={30},
  year={2017}
}

@inproceedings{hessel2021clipscore,
  title={{CLIPScore:} A Reference-free Evaluation Metric for Image Captioning},
  author={Hessel, Jack and Holtzman, Ari and Forbes, Maxwell and Bras, Ronan Le and Choi, Yejin},
  booktitle={EMNLP},
  year={2021}
}

@inproceedings{ma2025hpsv3,
  title={Hpsv3: Towards wide-spectrum human preference score},
  author={Ma, Yuhang and Wu, Xiaoshi and Sun, Keqiang and Li, Hongsheng},
  booktitle={ICCV},
  pages={15086--15095},
  year={2025}
}

@article{kolors,
  title={Kolors: Effective Training of Diffusion Model for Photorealistic Text-to-Image Synthesis},
  author={Kolors Team},
  journal={arXiv preprint},
  year={2024}
}

@inproceedings{chen2024pixart,
  title={Pixart-$\sigma$: Weak-to-strong training of diffusion transformer for 4k text-to-image generation},
  author={Chen, Junsong and Ge, Chongjian and Xie, Enze and Wu, Yue and Yao, Lewei and Ren, Xiaozhe and Wang, Zhongdao and Luo, Ping and Lu, Huchuan and Li, Zhenguo},
  booktitle={ECCV},
  pages={74--91},
  year={2024},
  organization={Springer}
}

@article{podell2023sdxl,
  title={Sdxl: Improving latent diffusion models for high-resolution image synthesis},
  author={Podell, Dustin and English, Zion and Lacey, Kyle and Blattmann, Andreas and Dockhorn, Tim and M{\"u}ller, Jonas and Penna, Joe and Rombach, Robin},
  journal={arXiv preprint arXiv:2307.01952},
  year={2023}
}

@article{xu2023imagereward,
  title={Imagereward: Learning and evaluating human preferences for text-to-image generation},
  author={Xu, Jiazheng and Liu, Xiao and Wu, Yuchen and Tong, Yuxuan and Li, Qinkai and Ding, Ming and Tang, Jie and Dong, Yuxiao},
  journal={NIPS},
  volume={36},
  pages={15903--15935},
  year={2023}
}

@inproceedings{kirstain2023pick,
  title={Pick-a-Pic: An Open Dataset of User Preferences for Text-to-Image Generation},
  author={Kirstain, Yuval and Polyak, Adam and Singer, Uriel and Matiana, Shahbuland and Penna, Joe and Levy, Omer},
  booktitle={NIPS},
  year={2023},
  doi={10.48550/arxiv.2305.01569}
}

@inproceedings{zhang2024learning,
  title={Learning multi-dimensional human preference for text-to-image generation},
  author={Zhang, Sixian and Wang, Bohan and Wu, Junqiang and Li, Yan and Gao, Tingting and Zhang, Di and Wang, Zhongyuan},
  booktitle={CVPR},
  pages={8018--8027},
  year={2024}
}

@article{saharia2022photorealistic,
  title={Photorealistic text-to-image diffusion models with deep language understanding},
  author={Saharia, Chitwan and Chan, William and Saxena, Saurabh and Li, Lala and Whang, Jay and Denton, Emily L and Ghasemipour, Kamyar and Gontijo Lopes, Raphael and Karagol Ayan, Burcu and Salimans, Tim and others},
  journal={NIPS},
  volume={35},
  pages={36479--36494},
  year={2022}
}

@article{fan2024fluid,
  title={Fluid: Scaling autoregressive text-to-image generative models with continuous tokens},
  author={Fan, Lijie and Li, Tianhong and Qin, Siyang and Li, Yuanzhen and Sun, Chen and Rubinstein, Michael and Sun, Deqing and He, Kaiming and Tian, Yonglong},
  journal={arXiv preprint arXiv:2410.13863},
  year={2024}
}

@misc{flux2024,
    author={Black Forest Labs},
    title={FLUX},
    year={2024},
}

@article{ho2022classifier,
  title={Classifier-free diffusion guidance},
  author={Ho, Jonathan and Salimans, Tim},
  journal={arXiv preprint arXiv:2207.12598},
  year={2022}
}

@inproceedings{sadat2024eliminating,
  title={Eliminating oversaturation and artifacts of high guidance scales in diffusion models},
  author={Sadat, Seyedmorteza and Hilliges, Otmar and Weber, Romann M},
  booktitle={ICLR},
  year={2024}
}

@article{song2025rethinking,
  title={Rethinking Oversaturation in Classifier-Free Guidance via Low Frequency},
  author={Song, Kaiyu and Lai, Hanjiang},
  journal={arXiv preprint arXiv:2506.21452},
  year={2025}
}

@article{hartwig2025survey,
  title={A survey on quality metrics for text-to-image generation},
  author={Hartwig, Sebastian and Engel, Dominik and Sick, Leon and Kniesel, Hannah and Payer, Tristan and Poonam, Poonam and Glockler, Michael and Bauerle, Alex and Ropinski, Timo},
  journal={TVCG},
  year={2025},
  publisher={IEEE}
}

@inproceedings{wu2023human,
  title={Human preference score: Better aligning text-to-image models with human preference},
  author={Wu, Xiaoshi and Sun, Keqiang and Zhu, Feng and Zhao, Rui and Li, Hongsheng},
  booktitle={ICCV},
  pages={2096--2105},
  year={2023}
}

@article{wu2023human2,
  title={Human preference score v2: A solid benchmark for evaluating human preferences of text-to-image synthesis},
  author={Wu, Xiaoshi and Hao, Yiming and Sun, Keqiang and Chen, Yixiong and Zhu, Feng and Zhao, Rui and Li, Hongsheng},
  journal={arXiv preprint arXiv:2306.09341},
  year={2023}
}

@article{shen2025directly,
  title={Directly aligning the full diffusion trajectory with fine-grained human preference},
  author={Shen, Xiangwei and Li, Zhimin and Yang, Zhantao and Zhang, Shiyi and Zhang, Yingfang and Li, Donghao and Wang, Chunyu and Lu, Qinglin and Tang, Yansong},
  journal={arXiv preprint arXiv:2509.06942},
  year={2025}
}

@article{sun2024autoregressive,
  title={Autoregressive model beats diffusion: Llama for scalable image generation},
  author={Sun, Peize and Jiang, Yi and Chen, Shoufa and Zhang, Shilong and Peng, Bingyue and Luo, Ping and Yuan, Zehuan},
  journal={arXiv preprint arXiv:2406.06525},
  year={2024}
}

@inproceedings{wang2023exploring,
  title={Exploring clip for assessing the look and feel of images},
  author={Wang, Jianyi and Chan, Kelvin CK and Loy, Chen Change},
  booktitle={AAAI},
  volume={37},
  number={2},
  pages={2555--2563},
  year={2023}
}

@article{qwen2.5,
  title={Qwen2.5 Technical Report},
  author={Bai, Jinze and others},
  journal={arXiv preprint arXiv:2409.12191},
  year={2024}
}

@article{wang2024qwen2,
  title={Qwen2-vl: Enhancing vision-language model's perception of the world at any resolution},
  author={Wang, Peng and Bai, Shuai and Tan, Sinan and Wang, Shijie and Fan, Zhihao and Bai, Jinze and Chen, Keqin and Liu, Xuejing and Wang, Jialin and Ge, Wenbin and others},
  journal={arXiv preprint arXiv:2409.12191},
  year={2024}
}

@misc{wu2025qwenimagetechnicalreport,
      title={Qwen-Image Technical Report}, 
      author={Chenfei Wu and Jiahao Li and Jingren Zhou and Junyang Lin and Kaiyuan Gao and Kun Yan and Sheng-ming Yin and Shuai Bai and Xiao Xu and Yilei Chen and Yuxiang Chen and Zecheng Tang and Zekai Zhang and Zhengyi Wang and An Yang and Bowen Yu and Chen Cheng and Dayiheng Liu and Deqing Li and Hang Zhang and Hao Meng and Hu Wei and Jingyuan Ni and Kai Chen and Kuan Cao and Liang Peng and Lin Qu and Minggang Wu and Peng Wang and Shuting Yu and Tingkun Wen and Wensen Feng and Xiaoxiao Xu and Yi Wang and Yichang Zhang and Yongqiang Zhu and Yujia Wu and Yuxuan Cai and Zenan Liu},
      year={2025},
      eprint={2508.02324},
      archivePrefix={arXiv},
      primaryClass={cs.CV},
      url={https://arxiv.org/abs/2508.02324}, 
}

@misc{li2024playground,
      title={Playground v2.5: Three Insights towards Enhancing Aesthetic Quality in Text-to-Image Generation}, 
      author={Daiqing Li and Aleks Kamko and Ehsan Akhgari and Ali Sabet and Linmiao Xu and Suhail Doshi},
      year={2024},
      eprint={2402.17245},
      archivePrefix={arXiv},
      primaryClass={cs.CV}
}

@article{li2024hunyuan,
  title={Hunyuan-dit: A powerful multi-resolution diffusion transformer with fine-grained chinese understanding},
  author={Li, Zhimin and Zhang, Jianwei and Lin, Qin and Xiong, Jiangfeng and Long, Yanxin and Deng, Xinchi and Zhang, Yingfang and Liu, Xingchao and Huang, Minbin and Xiao, Zedong and others},
  journal={arXiv preprint arXiv:2405.08748},
  year={2024}
}

@inproceedings{ke2021musiq,
  title={Musiq: Multi-scale image quality transformer},
  author={Ke, Junjie and Wang, Qifei and Wang, Yilin and Milanfar, Peyman and Yang, Feng},
  booktitle={ICCV},
  pages={5148--5157},
  year={2021}
}

@inproceedings{yang2022maniqa,
  title={Maniqa: Multi-dimension attention network for no-reference image quality assessment},
  author={Yang, Sidi and Wu, Tianhe and Shi, Shuwei and Lao, Shanshan and Gong, Yuan and Cao, Mingdeng and Wang, Jiahao and Yang, Yujiu},
  booktitle={CVPR},
  pages={1191--1200},
  year={2022}
}

@inproceedings{zhang2023blind,
  title={Blind image quality assessment via vision-language correspondence: A multitask learning perspective},
  author={Zhang, Weixia and Zhai, Guangtao and Wei, Ying and Yang, Xiaokang and Ma, Kede},
  booktitle={CVPR},
  pages={14071--14081},
  year={2023}
}

@article{ding2021cogview,
  title={Cogview: Mastering text-to-image generation via transformers},
  author={Ding, Ming and Yang, Zhuoyi and Hong, Wenyi and Zheng, Wendi and Zhou, Chang and Yin, Da and Lin, Junyang and Zou, Xu and Shao, Zhou and Yang, Hongxia and others},
  journal={NIPS},
  volume={34},
  pages={19822--19835},
  year={2021}
}

@inproceedings{esser2024scaling,
  title={Scaling rectified flow transformers for high-resolution image synthesis},
  author={Esser, Patrick and Kulal, Sumith and Blattmann, Andreas and Entezari, Rahim and M{\"u}ller, Jonas and Saini, Harry and Levi, Yam and Lorenz, Dominik and Sauer, Axel and Boesel, Frederic and others},
  booktitle={ICML},
  year={2024}
}

@inproceedings{rombach2022high,
  title     = {High-Resolution Image Synthesis with Latent Diffusion Models},
  author    = {Rombach, Robin and Blattmann, Andreas and Lorenz, Dominik and Esser, Patrick and Ommer, Bj{\"o}rn},
  booktitle = {CVPR},
  year      = {2022},
  pages     = {10684--10695}
}

@inproceedings{chang2022maskgit,
  title     = {MaskGIT: Masked Generative Image Transformer},
  author    = {Chang, Huiwen and Zhang, Han and Jiang, Lu and Liu, Ce and Freeman, William T. and Rubinstein, Michael and Zhang, Dingdong and Su, Li and Essa, Irfan and Hoppe, Hugues},
  booktitle = {CVPR},
  year      = {2022}
}

@article{esser2023structure,
  title   = {Structure and Content-Guided Video Synthesis with Autoregressive Transformers},
  author  = {Esser, Patrick and Blattmann, Andreas and Rombach, Robin},
  journal = {arXiv preprint arXiv:2305.13455},
  year    = {2023}
}

@article{peebles2023dit,
  title   = {Scalable Diffusion Models with Transformers},
  author  = {Peebles, William and Xie, Saining},
  journal = {arXiv preprint arXiv:2212.09748},
  year    = {2023},
  note    = {Introduces DiT architecture}
}

@article{esser2024sd3,
  title   = {Scaling Rectified Flow Transformers for High-Resolution Image Synthesis},
  author  = {Esser, Patrick and Rombach, Robin and Blattmann, Andreas and Savinov, Nikolay and Koh, Jing Yu Chao and Zhang, Yujun and and others},
  journal = {arXiv preprint arXiv:2403.03206},
  year    = {2024},
  note    = {Stable Diffusion 3}
}
